\documentclass[letterpaper, 10 pt, conference]{ieeeconf}  % Comment this line out if you need a4paper

\IEEEoverridecommandlockouts                              % This command is only needed if 
\usepackage{graphics} 
\usepackage{epsfig} 
\usepackage{mathptmx} 
\usepackage{times}
\usepackage{amsmath} 
\usepackage{amssymb} 
\usepackage{amsfonts}
\usepackage{booktabs}
\usepackage{textcomp}

\title{\LARGE \bf
Bundling and Tumbling in Bacterial-inspired Bi-flagellated Soft Robots for Attitude Adjustment
}

\author{Zhuonan Hao$^{1}$, Siddharth Zalavadia$^{1}$, Mohammad Khalid Jawed$^{1}$
\thanks{}
\thanks{$^{1}$Department of Mechanical \& Aerospace Engineering, University of California, Los Angeles, 420 Westwood Plaza, Los Angeles, CA 90095}}

\begin{document}

\maketitle
\thispagestyle{empty}
\pagestyle{empty}

%%%%%%%%%%%%%%%%%%%%%%%%%%%%%%%%%%%%%%%%%%%%%%%%%%%%%%%%%%%%%%%%%%%%%%%%%%%
%%%%%%%%%%%%%%%%%%%%%%%%%%%   Section Divider   %%%%%%%%%%%%%%%%%%%%%%%%%%%
%%%%%%%%%%%%%%%%%%%%%%%%%%%%%%%%%%%%%%%%%%%%%%%%%%%%%%%%%%%%%%%%%%%%%%%%%%%

\begin{abstract}

We create a mechanism inspired by bacterial swimmers, featuring two flexible flagella with individual control over rotation speed and direction in viscous fluid environments. Using readily available materials, we design and fabricate silicone-based helical flagella. To simulate the robot's motion, we develop a physics-based computational tool, drawing inspiration from computer graphics. The framework incorporates the Discrete Elastic Rod method, modeling the flagella as Kirchhoff's elastic rods, and couples it with the Regularized Stokeslet Segments method for hydrodynamics, along with the Implicit Contact Model to handle contact. This approach effectively captures polymorphic phenomena like bundling and tumbling. Our study reveals how these emergent behaviors affect the robot's attitude angles, demonstrating its ability to self-reorient in both simulations and experiments. We anticipate that this framework will enhance our understanding of the directional change capabilities of flagellated robots, potentially stimulating further exploration on microscopic robot mobility.
\end{abstract}

%%%%%%%%%%%%%%%%%%%%%%%%%%%%%%%%%%%%%%%%%%%%%%%%%%%%%%%%%%%%%%%%%%%%%%%%%%%%%%%%
%%%%%%%%%%%%%%%%%%%%%%%%%%%%%%%%%%%%%%%%%%%%%%%%%%%%%%%%%%%%%%%%%%%%%%%%%%%
%%%%%%%%%%%%%%%%%%%%%%%%%%%   Section Divider   %%%%%%%%%%%%%%%%%%%%%%%%%%%
%%%%%%%%%%%%%%%%%%%%%%%%%%%%%%%%%%%%%%%%%%%%%%%%%%%%%%%%%%%%%%%%%%%%%%%%%%%
\section{Introduction}
\label{intro}

Bacterial locomotion has long stood as a captivating frontier in the realm of robotics and bio-inspired engineering ~\cite{brennen1977fluid}. The design and control of highly maneuverable robotic systems have been informed by the propulsion mechanisms observed in nature, especially those employed by motile bacteria \cite{berg2000motile}. These microorganisms, often equipped with slender helical appendages protruding from their cells, exhibit remarkable agility in their aquatic environments \cite{sowa2008bacterial, leifson1960atlas}. Drawing inspiration from the natural world's efficiency and adaptability, roboticists have sought to emulate these bacterial propulsion methods with single \cite{du2021simple, armanini2021flagellate, magdanz2013development} or multiple flagella \cite{ye2013rotating, lim2023bacteria, hao2023modeling}. Multi-flagellated swimmer have mainly two modes of locomotion, i.e., run and tumble, via the intricate interplay of multiple slender flagella \cite{berg2003rotary}. However, current research primarily emphasizes the running ability of multi-flagellated robots, with less attention directed toward the realization of direction-changing mechanisms.

Recognizing this gap, we delve into the realm of attitude adjustment for multi-flagellated robots. The slender structure of these robots exhibits intriguing polymorphic transitions, including buckling, bundling, and tumbling \cite{flores2005study, vogel2013rotation, son2013bacteria, jawed2015propulsion, dvoriashyna2021hydrodynamics}, setting them apart from traditional rigid robots. Their soft, deformable flagella provide them with remarkable agility and the ability to navigate intricate and confined spaces. However, the intricate nonlinear nature of their structure, coupled with the complexities of fluid and contact interactions, underscores the need for in-depth exploration. 

\begin{figure}[ht!]
    \centering
    \includegraphics[width=\linewidth]{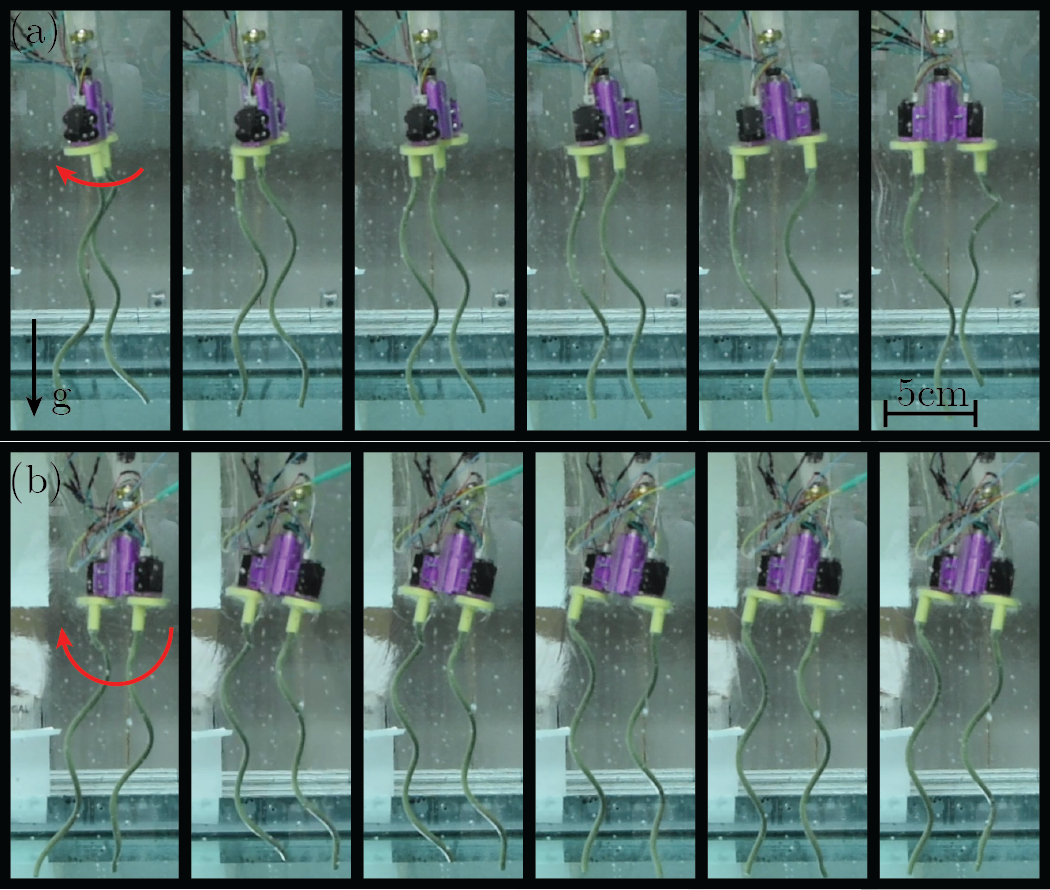}
    \caption{Attitude adjustment of bi-flagellated robot in the viscous fluid (glycerin). (a) Rotation in the horizontal plane (pointing opposite to gravity direction). (b) Rotation in the vertical plane (pointing out to the paper). }
    \label{fig:overview}
\end{figure}

Since it is difficult to systematically manipulate the physical parameters of microscopic natural bacteria, we choose an analog robotic system which can - in turn - inform the original natural system that motivated its design.This macroscopic bi-flagellated system is affixed to a ball joint, to facilitate the characterization and quantification of its attitude adjustment capabilities. As visualized in Figure \ref{fig:overview}, the robot is submerged in viscous fluid medium to faithfully emulate the environment bacteria thrive in. This robotic system exhibits the capacity to rotate along different body axes, i.e., yaw and pitch angle, reproducing the bacterial ability. Furthermore, to understand this phenomenon and explore the parameter space governing attitude adjustment, we introduce a comprehensive computational framework. This framework addresses critical aspects such as elasticity, long-range hydrodynamics, and the handling of physical contacts, each of which has demonstrated its efficacy in prior research \cite{tong2023fully, lim2023bacteria}. 

The paper is structured as follows. Section \ref{sec:experiment} provides a comprehensive overview of the experimental setup and robot structure, offering valuable insights into the practical aspects of our study. In Section \ref{sec:method}, we delve into the details of the computational framework, which integrates various methodologies, including the Discrete Elastic Rod (DER) \cite{bergou2008discrete}, Regularized Stokeslet Segments (RSS) \cite{cortez2018regularized}, and Implicit Contact Model (IMC) \cite{tong2023fully}. This section also addresses the specific considerations for modeling the rotational dynamics of the bi-flagellated robot. Moving on to Section \ref{sec:result}, we present a fundamental analysis of the attitude adjustment, encompassing critical aspects such as the reaction forces, the stability of Euler angles, and the range of attainable attitudes for the robot. This section delves into the core findings of our work, providing profound control application of the bi-flagellated robot. Lastly, in Section \ref{sec:conclusion}, we conclude our work by summarizing our key discoveries and outlining potential directions for future research in this exciting field.

%%%%%%%%%%%%%%%%%%%%%%%%%%%%%%%%%%%%%%%%%%%%%%%%%%%%%%%%%%%%%%%%%%%%%%%%%%%
%%%%%%%%%%%%%%%%%%%%%%%%%%%   Section Divider   %%%%%%%%%%%%%%%%%%%%%%%%%%%
%%%%%%%%%%%%%%%%%%%%%%%%%%%%%%%%%%%%%%%%%%%%%%%%%%%%%%%%%%%%%%%%%%%%%%%%%%%
\section{Experiment Setup}
\label{sec:experiment}

\subsection{Robotic structure}
\begin{figure}[ht!]
    \centering
    \includegraphics[width=\linewidth]{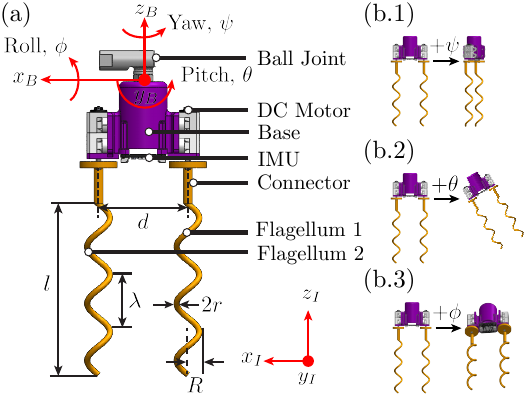}
    \caption{Robot schematic and attitude representation. (a) The bi-flagellated robot comprises an assembly base and two identical soft flagella. Each flagellum with length $l$, cross-sectional radius $r$, helix pitch $\lambda$, and helix radius $R$ is actuated by a miniature brushed DC motor. The robot is affixed to a ball joint to reorient. (b) Yaw angle $\psi$ (b.1), Pitch angle $\theta$ (b.2), and Roll angle $\phi$ (b.3) signifies the angular displacement around the vertical axis $z_B$, lateral axis $y_B$, and longitudinal axis $x_B$, respectively. }
    \label{fig:model}
\end{figure}

The robot illustrated in Figure \ref{fig:model}(a) consists of an assembly base and two helical flagella connected to motor shafts. The base, with mass $m$ of $94$ g, features a central cylinder with a radius $r_h$ of $13.5$ mm and a height $h$ of $37.5$ mm. Two small brushed DC motors, rated at $6$ V voltage and capable of a stall current of $1.5$ A, are positioned on opposite sides of the cylinder at a distance $d$ of $45$ mm. The motors allow for control of rotation direction and speed through pulse-width modulation (PWM) signals from a microcontroller (Arduino Nano). Additionally, the robot is equipped with an inertial measurement unit (IMU) (MPU6050) located on the bottom plane of the base. This IMU tracks the robot's orientation in terms of Euler angles, including yaw $\psi$, pitch $\theta$, and roll $\phi$. These angles are defined in the body frame $x_B-y_B-z_B$ and are illustrated in Figure \ref{fig:model}(b).

\begin{figure}[ht!]
    \centering
    \includegraphics[width=\linewidth]{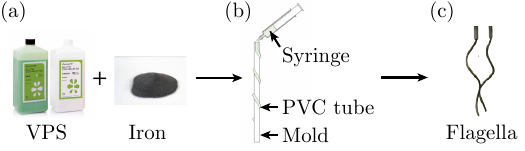}
    \caption{Fabrication of a soft flagellum. (a) Step 1: blending VPS with iron powder to precisely calibrate the density of the silicone composite. (b) Step 2: using a syringe to inject the mixture into the PVC tube preformed by a mold. (c) Step 3: slitting the tube to extract the silicone elastomer. }
    \label{fig:manufacturing}
\end{figure}

Referring to Figure \ref{fig:manufacturing}, the soft flagella are fabricated using VPS material (Elite Double 32). To modulate the silicone elastomer's density, the base and catalyst are blended with iron powder. The resulting mixture is then injected into pre-shaped PVC tubes, and after tube slitting, the soft flagellum is assembled onto the robot base. These helical flagella have specific dimensions, including a fixed cross-sectional radius $r$ of $6$ mm, a helix radius $R$ of $8.89$ mm, a helix pitch $\lambda$ of $76$ mm, and a helix axial length $l$ of $196.2$ mm. The silicone composite exhibits a slender profile, possessing a Young's modulus $E$ of $1.255\times10^6$ Pa, Poisson's ration $\nu$ of 0.5, and density $\rho$ of $1450$ kg/m$^3$. Each flagellum is actuated within a range of $0$ to $90$ rpm, ensuring that the Reynolds number $Re$ remains below $0.1$, as defined by the formula $Re = \rho \omega R r_0/\mu$.

% \begin{table}[ht!]
%   \centering
%   \caption{Parameters with symbolic representations}
%   \label{tab:para}
%   \begin{tabular}{ cccc }
%     \toprule
%     Symbol & Value & Unit & Description \\  
%     $r_m$ & -2 & mm & Mass center shift \\ 
%     $I_x$  & 1.63$\times10^{-5}$ & kg$\cdot$m$^2$ & Moment of inertia around the x-axis  \\ 
%     $I_y$  & 1.63$\times10^{-5}$ & kg$\cdot$m$^2$ & Moment of inertia around the y-axis  \\ 
%     $I_z$  & 0.91$\times10^{-5}$ & kg$\cdot$m$^2$ & Moment of inertia around the z-axis  \\ 
%     \midrule
%     $C_r$ & 2.3 & 1 & Rotational drag coefficient \\
%     \bottomrule
%   \end{tabular}
% \end{table}

\subsection{Experimental platform}
The experimental setup, as depicted in Figure \ref{fig:overview}, includes a glycerin tank measuring $122$ cm in length, $45$ cm in width, and $51.5$ cm in height. Glycerin, chosen for its density $\rho_{f}$ of $1.26$ g/ml and viscosity $\mu$ of $1$ Pa$\cdot$s at $25^\circ$ Celsius, serves as the surrounding viscous medium. To restrict the robot's motion, it is secured in place by a positioning frame featuring a ball joint that permits limited rotational movement. This setup ensures that the IMU located on the robot's central axis can accurately measure yaw, pitch, and roll angles over time. The IMU data is transmitted to an external microcontroller, responsible for both data processing and the independent control of the DC motors, enabling us to adjust their rotational speed and direction.

%%%%%%%%%%%%%%%%%%%%%%%%%%%%%%%%%%%%%%%%%%%%%%%%%%%%%%%%%%%%%%%%%%%%%%%%%%%
%%%%%%%%%%%%%%%%%%%%%%%%%%%   Section Divider   %%%%%%%%%%%%%%%%%%%%%%%%%%%
%%%%%%%%%%%%%%%%%%%%%%%%%%%%%%%%%%%%%%%%%%%%%%%%%%%%%%%%%%%%%%%%%%%%%%%%%%%
\section{Numerical method}
\label{sec:method}
The computational framework for a bi-flagellated system comprises three key components: (1) the Discrete Elastic Rods method for modeling the nonlinear structural elasticity, (2) the Regularized Stokeslet Segments method for handling long-range hydrodynamics, and (3) the Implicit Contact Model for realistic contact interactions. To simulate the dynamic behavior of the system, we incorporate all the reaction forces and moments generated by these methods into the governing equation of rotation. This comprehensive approach enables an effective modeling and analysis of the system's motion response.

\begin{figure}[ht!]
    \centering
    \includegraphics[width=\linewidth]{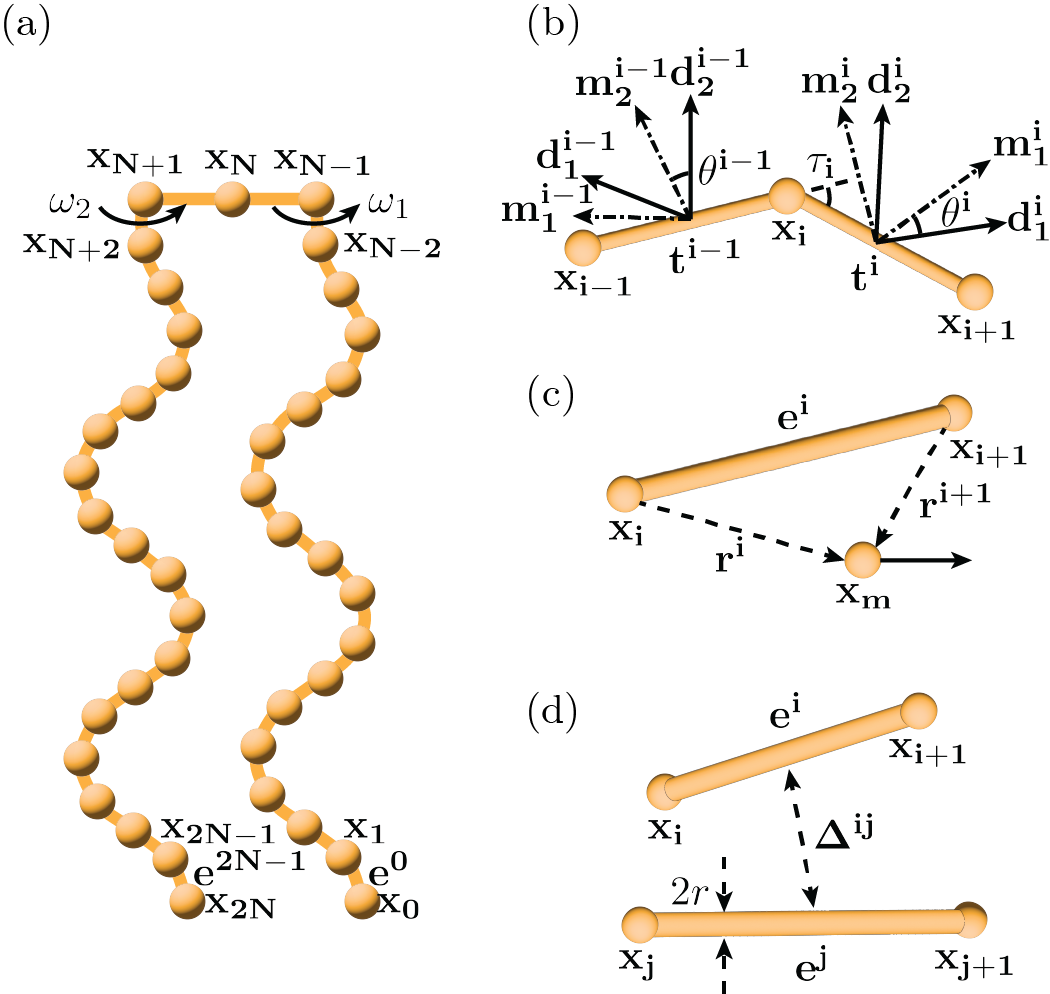}
    \caption{Computation framework of bi-flagellated robot. (a) Discretization scheme. Each soft flagellum is discretized into $\mathbf{N}$ nodes (denoted as $\mathbf{x_i}$) and $\mathbf{N-1}$ edges (denoted as $\mathbf{e^i}$), and connected to the base node $\mathbf{x_N}$. (b) Representation of material frame ([$\mathbf{m_{1}^{i-1}}, \mathbf{m_{2}^{i-1}}, \mathbf{t_{i-1}}$] and [$\mathbf{m_{1}^{i}}, \mathbf{m_{2}^{i}}, \mathbf{t_{i}}$]) and reference frame ([$\mathbf{d_{1}^{i-1}}, \mathbf{d_{2}^{i-1}}, \mathbf{t_{i-1}}$] and [$\mathbf{d_{1}^{i}}, \mathbf{d_{2}^{i}}, \mathbf{t_{i}}$]). The signed angle from $\mathbf{d_1^i}$ to $\mathbf{m_1^i}$ is $\mathbf{\theta^i}$ and twist at node $\mathbf{x_i}$ is $\mathbf{\tau_i} = \mathbf{\theta^i}-\mathbf{\theta^{i-1}}$. (c) Regularized Stokeslets Segments. This method establishes a connection between the velocity $\mathbf{\dot{x}_m}$ at a point $\mathbf{x_m}$ and the forces acting on the nodes $\mathbf{x_i}$ and $\mathbf{x_{i+1}}$. (d) Implicit contact model. The artificial contact energy is defined by the minimal distance $\Delta_{\mathbf{ij}}$ between two edges $\mathbf{e^i}$ and $\mathbf{e^j}$. }
    \label{fig:discrete}
\end{figure}

\subsection{Discrete elastic rods}
\label{sec:der}
Figure \ref{fig:discrete}(a) provides a discrete schematic of the bi-flagellated robot. In our modeling approach, each flagellum is represented using Kirchhoff's elastic rod, commonly employed in computer graphics. The DER algorithm discretizes each Kirchhoff rod into $\mathbf{N}$ nodes with the segment length $\Delta l$ of $5$ mm, denoted as ${\mathbf{x_0, x_1, ..., x_{N-1}}}$ and ${\mathbf{x_{N+1}, x_1, ..., x_{2N}}}$. To simulate the rotation dynamics of the robot, we introduce an additional node, $\mathbf{x_N}$, which represents the robot base and connects with the endpoints of the two rods. Notably, there is no relative motion among node $\mathbf{x_{N-1}}$, $\mathbf{x_{N+1}}$, and $\mathbf{x_N}$. Consequently, the system comprises a total of $\mathbf{2N+1}$ nodes (denoted as $\mathbf{n_x}$), characterized by $\mathbf{X} = \{\mathbf{x_0, x_1, ..., x_{2N}}\}$, corresponding to a total of $\mathbf{2N}$ edge vectors (denoted as $\mathbf{n_e}$), expressed as $\mathbf{E} = \{\mathbf{e^0, e^1, ..., e^{2N-1}}\}$, with $\mathbf{e^i = x_{i+1} - x_{i}}$.

As depicted in Figure \ref{fig:discrete}(b), each edge $\mathbf{e^i}$ is associated with two sets of orthonormal frames, responsible for accounting for rotation. These frames include a reference frame, denoted as $\{\mathbf{{d^i_1, d^i_2, t^i}}\}$, and a material frame, expressed as $\{\mathbf{{m^i_1, m^i_2, t^i}}\}$. Notably, both frames share the tangent vector $\mathbf{t^i = e^i/|e^i|}$ as one of their directors. The reference frame is initialized at time $t=0$ and is updated at each time step through time-parallel transport. In contrast, the material frame can be calculated based on a scalar twist angle, $\theta^{{\mathbf{i}}}$.

Consequently, for each flagellum (in this case, we focus on flagellum 1 for simplicity), the spatial positions of nodes, $\mathbf{x_i} = (x_{\mathbf{i}},y_{\mathbf{i}},z_{\mathbf{i}})$, and the twist angle $\theta^{{\mathbf{i}}}$ collectively constitute the degrees of freedom (DOF) vector, which has a size of $\mathbf{4N-1}$, i.e., 
\begin{equation*}
    \mathbf{q} = [\mathbf{x_0},\mathbf{\theta^0},\mathbf{x_1},...,\mathbf{x_{N-2}},\mathbf{\theta^{N-2}},\mathbf{x_{N-1}}]^T,
\end{equation*}
where superscript $^T$ represent transpose. The equations of motion (EOM) of the \textit{i}-th DOF is
\begin{equation*}
    m_{\mathbf{i}}\Ddot{q}_{\mathbf{i}} + \frac{\partial E_{\text{potential}}}{\partial q_{\mathbf{i}}} + f^{\text{external}} = 0,
\end{equation*}
where $E_{\text{potential}}$ is formulated in the remainder of this part, $f^{\text{external}}$ consists of external hydrodynamics forces $\mathbf{f}^h$ and contact force $\mathbf{f}^c$ that are addressed in the Section \ref{sec:rss} and \ref{sec:imc}. Note that $m_{\mathbf{i}}$ stands for mass (unit: kg) if $q_{\mathbf{i}}$ corresponds to spatial position and moment of inertia (unit: kg-m$^2$) if $q_{\mathbf{i}}$ corresponds to twist angle.

According to Kirchhoff's rod theory, the potential energy can be assumed to be a sum of elastic stretching, bending, and twisting energies, i.e.,
\begin{equation*}
\begin{aligned}
    E_{\text{potential}} &= E^{s} + E^{b} + E^{t}\\ 
    &= \sum ^{\mathbf{N-2}}_{\mathbf{i=0}} E^{s}_\mathbf{k} + \sum ^{\mathbf{N-2}}_{\mathbf{i=1}} E^{b}_\mathbf{k} + \sum ^{\mathbf{N-2}}_{\mathbf{i=1}} E^{t}_\mathbf{k}.
\end{aligned}
\end{equation*}

\textbf{Stretching energy}. The axial stretch, $\epsilon$, of edge $\mathbf{e^i}$ is
\begin{equation*}
    \epsilon^{\mathbf{i}} = \mathbf{\frac{\|e^i\|}{\|\bar{e}^i\|}}-1,
\end{equation*}
where $\mathbf{\|\bar{e}^i\|}$ is length of the edge in undeformed state. The stretching energy along edge $\mathbf{e^k}$ is
\begin{equation*}
    E^{s}_\mathbf{i} = \frac{1}{2}EA (\epsilon^{\mathbf{i}})^2 \|\mathbf{\bar{e}^{i}}\|,
\end{equation*}
where $EA = E\pi r^2$ is the stretching stiffness, $E$ is the Young's modulus, and $r$ is the rod radius.

\textbf{Bending energy}. Bending strain is measured at each node $\mathbf{x_i}$ through the curvature binormal vector
\begin{equation*}
        (\kappa \mathbf{b})_{\mathbf{i}} = \mathbf{\frac{\text{2}e^{i-1}\times e^i}{\|e^{k-1}\|\|e^{i}\|+e^{i-1}\cdot e^{i}}}.
\end{equation*}

The scalar curvatures along the first and second material directors using the curvature binomial are 
\begin{equation*}
\begin{aligned}
    \kappa^{(1)}_{\mathbf{i}} = \frac{1}{2}(\mathbf{m^{i-1}_2}+\mathbf{m^{i}_2})\cdot (\kappa \mathbf{b})_{\mathbf{i}}, \\
    \kappa^{(2)}_{\mathbf{i}} = \frac{1}{2}(\mathbf{m^{i-1}_1}+\mathbf{m^{i}_1})\cdot (\kappa \mathbf{b})_\mathbf{i}.
\end{aligned}
\end{equation*}

The bending energy at node $\mathbf{x_k}$ is 
\begin{equation*}
    E^{b}_{\mathbf{i}} = \frac{1}{2} \frac{EI}{\Delta l_{\mathbf{i}}} [(\kappa^{(1)}_{\mathbf{i}} - \bar{\kappa}^{(1)}_{\mathbf{i}})^2 + (\kappa^{(2)}_{\mathbf{i}} - \bar{\kappa}^{(2)}_{\mathbf{i}})^2 ],
\end{equation*}
where $\Delta l_{\mathbf{i}} = \frac{1}{2} (\|\mathbf{\bar{e}^{i-1}}\| + \|\mathbf{\bar{e}^{i}}\|)$ is the Voronoi length for the $\mathbf{i}$-th node, $\bar{\kappa}^{(1)}_{\mathbf{i}}$ and $\bar{\kappa}^{(2)}_{\mathbf{i}}$ are the material curvatures in undeformed state, and $EI = E\pi r^4 /4$ is the bending stiffness.

\textbf{Twisting energy}.
The twist between two consecutive edges at node $\mathbf{x_i}$ is 
\begin{equation*}
    \tau_{\mathbf{i}} = \theta^{\mathbf{i}} - \theta^{\mathbf{i-1}} + \Delta m_{\mathbf{i},\text{ref}},
\end{equation*}
where $\Delta m_{\mathbf{i},\text{ref}}$ is the reference twist, which is the twist of the reference frame as it moves from the $\mathbf{(i-1)}$-th edge to the $\mathbf{i}$-th edge. The twisting energy is
\begin{equation*}
    E^{t}_{\mathbf{i}} = \frac{1}{2} \frac{GJ}{\Delta l_{\mathbf{i}}} (\tau_{\mathbf{i}} - \bar{\tau}_{\mathbf{i}})^2,
\end{equation*}
where $\bar{\tau}_{\mathbf{i}}$ is the twist along the centerline in the undeformed state, $GJ = G\pi r^2/2$ is the twisting stiffness, and $G$ is the shear modulus.

\subsection{Regularized Stokeslets Segments}
\label{sec:rss}

We employ the Regularized Stokeslets Segments method to model the long-range hydrodynamics interaction between two flagella and viscous fluids. This method establishes a linear relationship between the velocity vector $\mathbf{U}$ (size $3\mathbf{n_x}$) at the node set $\mathbf{X}$ and the hydrodynamic force vector $\mathbf{F}$ (size $3\mathbf{n_x}$) acting upon them. This relationship is defined by a geometry-associated matrix $\mathbf{A}$ (size $3\mathbf{n_x}\times3\mathbf{n_x}$), i.e.,
\begin{equation}
\label{equ:linear_relation}
    \mathbf{U} = \mathbf{A}\mathbf{F}.
\end{equation}

Referring to Figure \ref{fig:discrete}(c), the RSS method characterizes the flow arising from a singular point force. It establishes a connection between the velocity $\mathbf{\dot{x}_m}$ at a point $\mathbf{x_m}$ and the forces exerted by each node on the fluid, i.e.,
\begin{equation*}
8 \pi \mu \mathbf{\dot{x}_m} = \sum_{\mathbf{i=0}}^{\mathbf{N-2}}(\mathbf{A}_1^{\mathbf{i}} \mathbf{f}^h_{\mathbf{i}}+\mathbf{A}_2^{\mathbf{i}} \mathbf{f}^h_{\mathbf{i+1}}),
\end{equation*}
where $\mathbf{f}^h_{\mathbf{i}}$ (size $3$) signifies the force vector corresponding to the force applied by the $\mathbf{i}$-th node onto the fluid. This force is equal in magnitude but opposite in direction to the hydrodynamic force acting on the $\mathbf{i}$-th node. The construction of matrices $\mathbf{A}_1^{\mathbf{i}}$ and $\mathbf{A}_2^{\mathbf{i}}$ follows the procedures in \cite{hao2023modeling}.

Matrix $\mathbf{A}$ is then assembled from matrices $\mathbf{A}_1^{\mathbf{i}}$ and $\mathbf{A}_2^{\mathbf{i}}$ with proper rearrangement. At each time step, by calculating the positions $\mathbf{X}$ and velocities $\mathbf{U}$ using the DER method, we initially derive matrix $\mathbf{A}$. Subsequently, we employ the Equation \ref{equ:linear_relation} to evaluate the hydrodynamic forces at each node and compute the resultant forces of each flagellum, i.e.,
\begin{equation*}
    \mathbf{F^1} = -\sum_{\mathbf{i=0}}^{\mathbf{N-1}} \mathbf{f}^h_{\mathbf{i}}, \mathbf{F^2} = -\sum_{\mathbf{i=N+1}}^{\mathbf{2N}} \mathbf{f}^h_{\mathbf{i}}. 
\end{equation*}
Lastly, we apply $\mathbf{F^1} = [F^1_x,F^1_y,F^1_z]$ and $\mathbf{F^2} = [F^2_x,F^2_y,F^2_z]$ on node $\mathbf{x_{N-1}}$ and $\mathbf{x_{N+1}}$ to study the rotation dynamics in Section \ref{sec:dynamics}.

\subsection{Implicit contact model}
\label{sec:imc}

In light of the physical rod-on-rod interaction between two flagella, we introduce the Implicit Contact Model to handle the contact interactions between the impenetrable edges. IMC is a fully implicit penalty-based frictional contact method that has demonstrated its capability to accurately capture challenging contact scenarios, including instances like flagella bundling. In contrast to previous studies \cite{lim2023bacteria}, which employed an adaptive contact model (Euler forward) in conjunction with DER, we have opted to replace this explicit approach with IMC for two compelling reasons: (i) The implicit (Euler backward) implementation of IMC can effectively circumvent numerical convergence issues, enhancing the stability and reliability of the simulation. (ii) The model's capacity to accommodate larger time steps without compromising accuracy greatly accelerates the simulation, contributing to computational efficiency and a more streamlined modeling process.

Referring to Figure \ref{fig:discrete}(d), we examine an edge-to-edge contact pair denoted as $\mathbf{x_{ij} := (x_i, x_{i+1}, x_j, x_{j+1})}$ from the node set $\mathbf{X}$. Detailed in \cite{tong2023fully}, the formulation of frictional contact forces follows the subsequent steps: (i) computing the minimal distance $\Delta^{\mathbf{ij}}$ between two edges $\mathbf{e^i}$ and $\mathbf{e^j}$, (ii) assessing the contact energy $E(\Delta^{\mathbf{ij}}, \delta)$ with respect to the rod cross-sectional radius $r$ and contact distance tolerance $\delta$ of $0.01$ mm, and (iii) determining the contact forces as the negative gradient of the contact energy, i.e.,
\begin{equation*}
    \mathbf{f}^c_{\mathbf{ij}} = -k \nabla_{\mathbf{x}} E(\Delta^{\mathbf{ij}}, \delta),
\end{equation*}
where $k$ is the contact stiffness to ensure the non-penetration. 

\subsection{Rotation dynamics}
\label{sec:dynamics}

Building upon the previously outlined computational framework, we delve into the development of the rotational dynamics for the bi-flagellated robot, which is attached to a ball joint. The primary objective of this analysis is to explore the potential for adjusting the robot's attitude. To mitigate the common singularity issues associated with Euler angles, we opt for quaternions represented as $\mathbf{q_o} = [q_w, q_x, q_y, q_z] \in \mathbb{R}^4$ and angular velocity components along the body axes $x_B, y_B, z_B$, denoted as $\boldsymbol{\omega_o} = [\omega_x, \omega_y, \omega_z] \in \mathbb{R}^3$, to describe the robot's orientation. If needed, the Euler angles $[\psi, \theta, \phi]$ can be easily converted by quaternions, e.g., \texttt{quat2eul} function available in MATLAB. The governing equation for rotation is articulated as follows
\begin{equation}
\label{equ:dynamics}
\begin{aligned}
    \frac{d\mathbf{q_o}}{dt} &= \frac{1}{2} \mathbf{q_o} \otimes
    \begin{bmatrix}
    0 \\
    \boldsymbol{\omega_o}
    \end{bmatrix}, \\
    \frac{d\boldsymbol{\omega_o}}{dt} &= \mathbf{J}^{-1}(\mathbf{M} - \boldsymbol{\omega_o}\times(\mathbf{J}\boldsymbol{\omega_o})),
\end{aligned}
\end{equation}
where $\mathbf{J}$ represents the moment of inertia matrix, $\otimes$ signifies quaternion multiplication, and $\mathbf{M} = [M_x, M_y, M_z]$ corresponds to the external torque applied to the robot's base along the $x_B$, $y_B, z_B$ axes.

Taking into account the impact of hydrodynamics and inertia, we derive the expression for the external torque as
\begin{equation*}
\begin{aligned}
    M_x &= (F^1_x - F^2_x)\times 0 + M_{hx}, \\
    M_y &= (F^1_z - F^2_z)\times\frac{d}{2} + M_{r} + M_{hy},\\
    M_z &= (F^1_y - F^2_y)\times\frac{d}{2} + M_{hz},
\end{aligned}
\end{equation*}
where $F^1_x, F^2_x, F^1_z, F^2_z, F^1_y, F^2_y$ are the flagella forces as described in Section \ref{sec:rss}. It's worth noting that although $F^1_x, F^2_x$ have nontrivial values, they do not effectively apply a torque to the robot since the lever arm is zero. Furthermore, $M_r$ represents the righting moment, which seeks to restore a tilted robot to an upright position due to the mass center shift $r_m$ measuring $2$ mm, with expressioin
\begin{equation}
\label{equ:righting}
    M_r = m g r_m \sin\theta,
\end{equation}
where $m$ is the base mass, $\theta$ is the titled pitch angle, and $g$ is gravitational acceleration. 

Lastly, we consider the viscous drag exerted on the robot base due to the angular velocity $\boldsymbol{\omega_o}$, i.e.,
\begin{equation*}
\begin{aligned}
    M_{hx} &= -2\pi\mu C_{rx} r_h^3\omega_x, \\
    M_{hy} &= -2\pi\mu C_{ry} r_h^3\omega_y,\\
    M_{hz} &= -2\pi\mu C_{rz} r_h^3\omega_z,
\end{aligned}
\end{equation*}
where $C_{rx},C_{ry},C_{rz}$ accounts for the nonsphericity of the robot's base and can be empirically determined, $\mu$ is the viscosity, and $r_h$ is the base radius.

%%%%%%%%%%%%%%%%%%%%%%%%%%%%%%%%%%%%%%%%%%%%%%%%%%%%%%%%%%%%%%%%%%%%%%%%%%%
%%%%%%%%%%%%%%%%%%%%%%%%%%%   Section Divider   %%%%%%%%%%%%%%%%%%%%%%%%%%%
%%%%%%%%%%%%%%%%%%%%%%%%%%%%%%%%%%%%%%%%%%%%%%%%%%%%%%%%%%%%%%%%%%%%%%%%%%%
\section{Results and discussion}
\label{sec:result}

\subsection{Analysis of force reaction}
\begin{figure}[ht!]
    \centering
    \includegraphics[width=\linewidth]{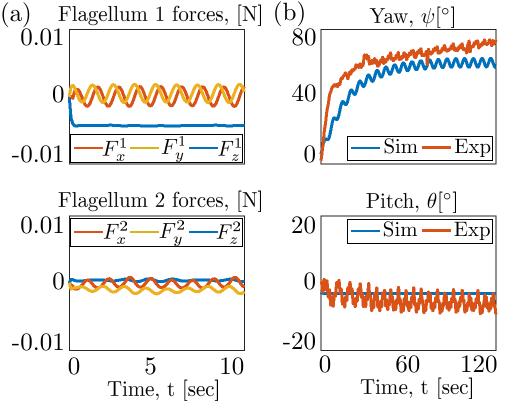}
    \caption{Time evolution of reaction forces and attitude angles under dual actuation of $\omega_1 = 50$ rpm and $\omega_2 = 10$ rpm. (a) Examination of Flagellum 1 (Top) and Flagellum 2 (Bottom) reveals that the vertical force component, $F_z$, rapidly stabilizes after an initial transition, while the horizontal forces, $F_x$ and $F_y$, exhibit sinusoidal behavior. For a right-handed helix, the counter-clockwise rotation ($\omega > 0$) generates an downward propulsion force. (b) Yaw angle $\psi$ and pitch angle $\theta$ eventually stabilize, reaching steady-state values of $\psi_{\text{ss}} = 60.76 ^\circ$ and $\theta_{\text{ss}} = -3.29 ^\circ$  in simulation, and $\psi_{\text{ss}} = 67.95 ^\circ$ and $\theta_{\text{ss}} = -4.96 ^\circ$ in experiment.
    }
    \label{fig:compEvo}
\end{figure}

The interaction between the flagella and the surrounding fluid medium leads to a distinct force reaction, illustrated in Figure \ref{fig:compEvo}(a). The flagella exhibit not only a stable propulsion thrust $F_z$ (blue curves), but also induce periodic forces $F_x$ and $F_y$ (orange and yellow curves, respectively).

The propulsion thrust $F_z^1$ and $F_z^2$ are generated by the two flagella at a distance $d$. This thrust gives rise to a torque that effectively changes the pitch angle $\theta$ within the $x_B-z_B$ plane. Simultaneously, the periodic forces, $F_y^1$ and $F_y^2$, possessing different phases and magnitudes, exert a torque responsible for altering the yaw angle $\psi$ within the $x_B-y_B$ plane.

However, it is important to note that due to the points of reaction being situated along the $x_B$ axis, the forces $F_x^1$ and $F_x^2$ do not contribute to changes in the roll angle $\psi$ within the $y_B-z_B$ plane. As a result, we deduce that in the context of bi-flagellated robots, the effective adjustment of attitude angles is limited to yaw and pitch. To modify the roll angle, additional flagella placed off the longitudinal axis $x_B$ or alterations in the positioning angle between the two flagella would be required.

\subsection{Stabilization of attitude angle}
Using- Equation \ref{equ:dynamics}, we investigate the time-dependent evolution of yaw and pitch angle while actuating two flagella with different speeds. Figure \ref{fig:compEvo}(b) displays the response of Euler angles when the force input follows the pattern presented in Figure \ref{fig:compEvo}(a). Notably, the two angles exhibit a convergence to stable values of $60.76^\circ$ and $-3.29^\circ$, respectively, which closely align with experimental results. In this section, we delve into an analysis of the factors contributing to the convergence observed in two attitude angles.

For the pitch angle $\theta$, the impact of the righting moment $M_r$ in Equation \ref{equ:righting} becomes particularly evident. The propulsion thrust does not lead to endless changes in the pitch angle $\theta$, and instead, it contributes to the establishment of a stable equilibrium value associated with the 
system's intrinsic properties. Similarly, the evolution of the yaw angle $\psi$ experiences constraints due to the presence of viscous drag in the fluid. This drag force acts as a restraint, preventing the yaw angle from continuously increasing and thereby contributing to the overall stability of the system.

As a consequence of these underlying principles, we observe that, for a given actuation of the two flagella, the robot's attitude always undergoes a gradual transition to a steady state, denoted as $\psi_{\text{ss}}$ and $\theta_{\text{ss}}$. This intriguing evolution provide valuable insights into the dynamic behavior of the bi-flagellated robot under the specified actuation conditions.

\subsection{Visualization of attainable attitude}

With the established convergence of attitude angles, we gain confidence in the robot's ability to attain a stable attitude over time. This stable attitude is defined in terms of the actuation speeds, $\omega_1$ and $\omega_2$, providing insight into the attitude adjustment possibilities of the bi-flagellated robot.

In our simulation, we systematically vary $\omega_1$ from $-90$ rpm to $90$ rpm and $\omega_2$ from $0$ to $90$ rpm in $10$ rpm intervals, resulting in a total of $190$ trials. In the experimental setup, we randomly select speed pairs ($\omega_1, \omega_2$) from the same range, totaling $75$ trials. Each trial is conducted over an extended period, approximately $120$ seconds, to ensure the emergence of attitude stability. To visualize the results, we present the steady-state values $|\psi_{\text{ss}}|$ and $|\theta_{\text{ss}}|$ (where $|\cdot|$ denotes the absolute value) in a colormap, using actuation speeds $\omega_1$ and $\omega_2$ as the $x$ and $y$ axes, respectively.

\begin{figure}[ht!]
    \centering
    \includegraphics[width=\linewidth]{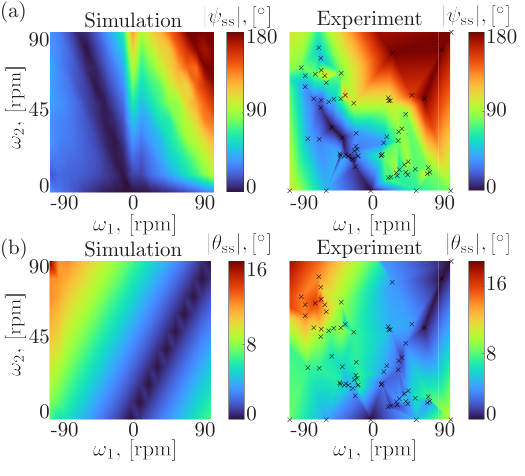}
    \caption{The steady-state values of (a) yaw angle $|\psi_{\text{ss}}|$ and (b) pitch angle $|\theta_{\text{ss}}|$ are depicted in response to  actuation speeds $\omega_1$ and $\omega_2$. The generated maps represent results from simulations (left) and experiments (right), based on 180 and 75 trials, respectively. $\times$ marks the experimental data points. }
    \label{fig:Direction}
\end{figure}

The results, depicted in Figure \ref{fig:Direction}, unveil the robot's remarkable capability to adjust its yaw and pitch angles through specific actuation configurations. For instance, to achieve the maximum yaw angle, one should explore the upper right corner of Figure \ref{fig:Direction}(a) with $|\psi_{\text{ss}}| = 180 ^\circ$ and $(\omega_1, \omega_2) = (+90, +90)$ rpm. Conversely, for maximum pitch angle, one should examine the upper left corner of Figure \ref{fig:Direction}(b) with $|\theta_{\text{ss}}| = 16 ^\circ$ and $(\omega_1, \omega_2) = (-90, +90)$ rpm. This information allows us to predict the expected attitude for any given actuation. For example, when $(\omega_1, \omega_2) = (0, +45)$ rpm, the attitude will stabilize at $|\psi_{\text{ss}}| = 56.22 ^\circ$ and $|\theta_{\text{ss}}| = 3.45^\circ$.

Notably, two distinctive lines become evident in both experimental and simulation results: $\omega_2 = -\omega_1$ in Figure \ref{fig:Direction}(a) and $\omega_2 = \omega_1$ in Figure \ref{fig:Direction}(b). Along these lines, the steady state value $|\psi_{\text{ss}}|$ or $|\theta_{\text{ss}}|$ reaches a local minimum in their respective neighborhoods. This intriguing behavior can be interpreted as follows. The minimal yaw angle region ($\omega_2 = -\omega_1$) arises from the opposing forces $F_y^1$ and $F_y^2$ generated by the two flagella, which effectively cancel each other out and prevent any significant rotation torque from being applied to the robot. Conversely, for minimal pitch angle region ($\omega_2 = \omega_1$), the thrust forces $F_z^1$ and $F_z^2$ remain balanced, resulting in no flip torque being applied to the robot. Consequently, the pitch angle remains largely unaffected. Therefore, to achieve simultaneous changes in both yaw and pitch angles, an unbalanced speed for the two flagella is necessary. 

In conclusion, the generated map not only illustrates all potential attitudes for the bi-flagellated robot but also serves as a valuable tool for pinpointing actuation speed. This map's applicability extends to various aspects of robot design, including flagellum geometry, elasticity, fluid viscosity, and robot structure. For clarity, we highlight four dimensionless variables influencing the map's pattern, including
\begin{equation*}
    \frac{\omega \mu l^4}{EI}, \frac{d}{l}, \frac{\lambda}{L}, \frac{R}{L}.
\end{equation*}

These insights provide a solid foundation for the efficient control of attitude  through strategic actuation configurations, with wide applications including untethered robots requiring precise reorientation.

%%%%%%%%%%%%%%%%%%%%%%%%%%%%%%%%%%%%%%%%%%%%%%%%%%%%%%%%%%%%%%%%%%%%%%%%%%%
%%%%%%%%%%%%%%%%%%%%%%%%%%%   Section Divider   %%%%%%%%%%%%%%%%%%%%%%%%%%%
%%%%%%%%%%%%%%%%%%%%%%%%%%%%%%%%%%%%%%%%%%%%%%%%%%%%%%%%%%%%%%%%%%%%%%%%%%%
\section{Conclusions and future work}
\label{sec:conclusion}

In summary, this study delves into the attitude adjustment capabilities of a scaled up bi-flagellated system. The investigation begins with an analysis of force reactions resulting from the flagella's interaction with the fluid, encompassing stable propulsion thrust and periodic forces. The research further explores the stabilization of the attitude angles, shedding light on the factors contributing to their convergence. A significant outcome is the visual representation of attainable attitudes in the generated map. Actuation speeds are systematically varied through simulations and experiments, illustrating the robot's capacity to fine-tune its yaw and pitch angles in response to specific actuation configurations.

The study emphasizes the practical applicability of this map for designing and controlling bi-flagellated robots, offering valuable insights into critical design parameters. Ultimately, we conclude these findings lay a strong foundation for precise attitude control of untethered robots.

%%%%%%%%%%%%%%%%%%%%%%%%%%%%%%%%%%%%%%%%%%%%%%%%%%%%%%%%%%%%%%%%%%%%%%%%%%%
%%%%%%%%%%%%%%%%%%%%%%%%%%%   Section Divider   %%%%%%%%%%%%%%%%%%%%%%%%%%%
%%%%%%%%%%%%%%%%%%%%%%%%%%%%%%%%%%%%%%%%%%%%%%%%%%%%%%%%%%%%%%%%%%%%%%%%%%%

\section{Acknowledge}
\label{sec:acknowledge}
We acknowledge financial support from the National Science Foundation under Grant numbers CAREER-2047663.

%%%%%%%%%%%%%%%%%%%%%%%%%%%%%%%%%%%%%%%%%%%%%%%%%%%%%%%%%%%%%%%%%%%%%%%%%%%
%%%%%%%%%%%%%%%%%%%%%%%%%%%   Section Divider   %%%%%%%%%%%%%%%%%%%%%%%%%%%
%%%%%%%%%%%%%%%%%%%%%%%%%%%%%%%%%%%%%%%%%%%%%%%%%%%%%%%%%%%%%%%%%%%%%%%%%%%

\bibliographystyle{ieeetr}
\bibliography{mybib}{}

\end{document}